\pdfoutput=1

\documentclass[11pt]{article}

\usepackage[final]{acl}

\usepackage{times}
\usepackage{latexsym}
\usepackage{graphicx}

\usepackage{booktabs}
\usepackage{multirow}
\usepackage{amssymb}
\usepackage{stfloats}

\usepackage[T1]{fontenc}

\usepackage[utf8]{inputenc}
\usepackage{url}

\usepackage{microtype}

\usepackage{inconsolata}

\title{Smotrom tvoja på ander drogoj verden! \\
Resurrecting Dead Pidgin with Generative Models: \\
Russenorsk Case Study}

\author{Alexey Tikhonov\\
  Independent researcher \\ Berlin, Germany \\
  \texttt{altsoph@gmail.com} \\
  \And
  Sergei Shteiner \\
  Independent researcher \\ Berlin, Germany \\
  \texttt{sergei.shteiner@gmail.com} \\
    \AND
  Anna Bykova \\
  Independent researcher \\ St. Petersburg, Russia \\
  \texttt{annakholmovaia@gmail.com} \\
     \And
  Ivan P. Yamshchikov \\
  CAIRO, THWS \\
  Würzburg, Germany \\
  \texttt{ivan.yamshchikov@thws.de} \\
}

\begin{document}
\maketitle
\begin{abstract}
Russenorsk, a pidgin language historically used in trade interactions between Russian and Norwegian speakers, represents a unique linguistic phenomenon. In this paper, we attempt to analyze its lexicon using modern large language models (LLMs)\footnote{We used anthropic/claude-3-5-sonnet-20241022 and openai/o1-2024-12-17}, based on surviving literary sources. We construct a structured dictionary of the language, grouped by synonyms and word origins. Subsequently, we use this dictionary to formulate hypotheses about the core principles of word formation and grammatical structure in Russenorsk and show which hypotheses generated by large language models correspond to the hypotheses previously proposed ones in the academic literature. We also develop a “reconstruction” translation agent that generates hypothetical Russenorsk renderings of contemporary Russian and Norwegian texts.
\end{abstract}

\section{Introduction}

This paper is part of a relatively recent trend to use Large Language Models (LLMs) for scientific discovery in various research areas, demonstrating that these natural language processing models have unprecedented generalization capacity. To our knowledge, this is the first result that demonstrates how large language models could be effectively used in a study of dead contact languages. We first provide a brief overview of LLM-driven scientific discovery and then dive deeper into Russenorsk and its properties.

\subsection{LLMs in Scientific Discovery}

Natural language processing methods could inform various research hypotheses in a very broad collection of research fields, be it biochemistry \newcite{ferruz2022protgpt2}, human psychology \newcite{safdari2023personality,sorokovikova2024llms}, or even ancient history \newcite{yamshchikov2022bert,riemenschneider2023exploring,gorovaia2024sui,SchmidtVY24}. Yet this trend is boosted with the arrival of the automated Scientific Discovery Agents \newcite{Jansen2024DISCOVERYWORLD} — AI-driven systems designed to autonomously conduct scientific research by generating hypotheses, designing and executing experiments, analyzing data, and formulating conclusions without human intervention. They are used in various fields, including mathematics \newcite{Lightman2023ProcessSupervision, Dong2023LLMScience}, social sciences \newcite{Yang2023LLMHypotheses}, astronomy \newcite{Nguyen2023AstroLLaMA, Ciuca2023AdversarialPrompting}, and chemistry \newcite{Jegorove2023BranchedFluorenylidene}. Notably, the 2024 Nobel Prize in Chemistry was awarded to David Baker, Demis Hassabis, and John Jumper for their groundbreaking work in computational protein design and protein structure prediction, highlighting the transformative impact of AI-driven methods in scientific discovery \newcite{Jumper2021AlphaFold}. Large Language Models (LLMs) are increasingly being used to support scientific research, particularly in data analysis and hypothesis generation. In our case, we explore how LLM-based Automated Scientific Discovery Agents can assist in studying Russenorsk and gaining new insights into this historical contact language.

\subsection{Russenorsk}

Russenorsk is a highly unusual pidgin. \newcite{Fuente_2020} states that its two primary lexifiers, Russian and Norwegian, contributed to its lexicon in roughly equal measure. This makes Russenorsk stand out from other pidgins. This statement is echoed in \newcite{stern2020russian}. \newcite{boldivzarova2021contemporary} concludes that "to understand the Russenorsk pidgin, some knowledge of the second lexifier is needed" and that "the Norwegian language skills are very beneficial but insufficient".
At the same time one has to admit that we have very limited amount of sources documenting the pidgin, and depending on the choice of the analyzed documents, different researchers come to drastically different conclusions about the properties of Russenorsk. For example, \newcite{kortlandt2000russenorsk} follows \newcite{brochjahr} and concludes that "Russenorsk is a variant of Norwegian with an admixture of Russian foreigners' talk and elements from the native language of the speaker". 

\subsection{Key Contributions}

This paper attempts to leverage large language models to analyze the structure of synonym formation in Russenorsk, as well as to understand the fundamental principles of phonetic and morphological transformations of words from Russian and Norwegian when incorporated into the Russenorsk lexicon.

We present three key contributions aimed at advancing the study of Russenorsk and its broader linguistic implications:
\begin{itemize}
    \item Expanded Russenorsk Vocabulary; we introduce a newly compiled set of lexemes that extends the existing Russenorsk wordlist, offering both contemporary scholars and digital tools a more comprehensive lexical resource. This vocabulary captures variant forms and rare lexical items noted in historical sources and field observations.
    \item Linguistic Discovery Assistant; building on this enriched lexicon, we propose a pipeline that automates critical steps in scientific inquiry—from dictionary-based feature extraction to the generation of hypotheses and preliminary translations. This framework enables researchers to rapidly test linguistic assumptions, compare usage across contexts, and refine their understanding of Russenorsk’s structural properties.
    \item Translation Reconstruction for Extinct Languages; finally, this paper is a case study presenting how the above pipeline can support the reconstruction of translations in a contact language no longer spoken. By leveraging parallel text alignment and informed lexical matching, our method offers a systematic approach to retrieving and interpreting content in Russenorsk, thus shedding light on its historical usage and guiding similar efforts for other extinct or endangered languages.
\end{itemize}

\section{New Russenorsk Vocabulary}
\label{sec:voc}

We present a new, extensive Russenorsk Vocabulary based on a variety of sources available in digital form. 

 To compile a Russenorsk vocabulary, we began by extracting lemmas from two primary sources:
the online repository at \url{https://en.wiktionary.org/wiki/Category:Russenorsk_lemmas}
and digitized version of \newcite{broch1984a}. This process yielded a CSV file with 691 entries. Each entry encompasses: 
\begin{enumerate} 
\item \textit{Russenorsk form} (unique words or fixed expressions, potentially with variant spellings), \item \textit{Cyrillic representation} (if identified), 
\item \textit{Translations in English/Norwegian/Russian}, 
\item \textit{Part-of-speech}, 
\item \textit{Comment} (additional notes or observations), 
\item \textit{Russenorsk example} from the source text, 
\item \textit{Norwegian translation} of the same example, 
\item \textit{Russian translation} of the Norwegian example (manually refined from automatic translations), 
\item \textit{Book page reference} (if an example was found), 
\item \textit{Link} to relevant Wiktionary information, 
\item \textit{IPA pronunciation} (if available). 
\end{enumerate}

After assembling the initial dictionary, we employed a pipeline (based on a Sonnet-driven system) to cluster entries with similar morphology, and subsequently grouped these by semantic proximity. This reformatting required manual review over several days to remove erroneous groupings and “hallucinated” entries. Finally, the cleaned dictionary was reviewed by additional collaborators, who discovered a few remaining mistakes that were subsequently corrected by hand.

\begin{figure}
    \centering
    \includegraphics[width=5cm,height=5cm]{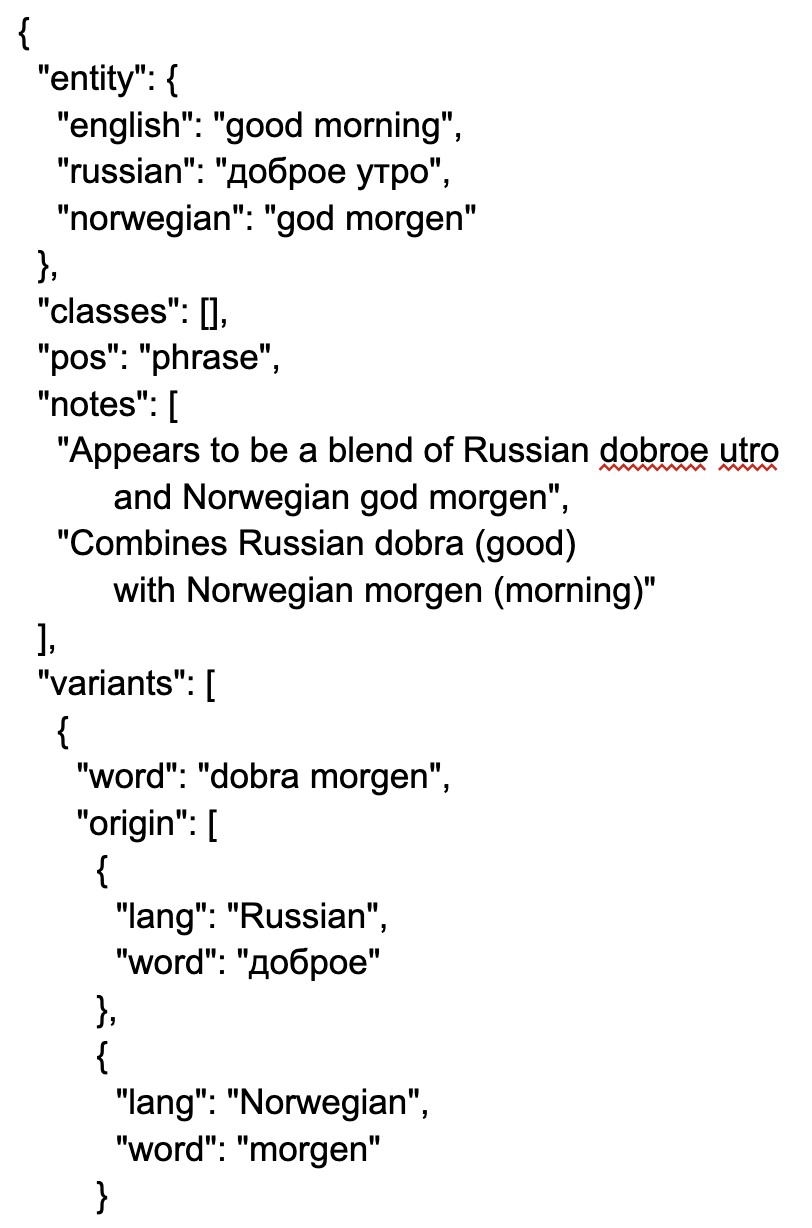}
    \caption{An example of a resulting vocabulary entry.}
    \label{fig:voc_example}
\end{figure}

Figure \ref{fig:voc_example} provides an example of a vocabulary entry. Full vocabulary is available online\footnote{\url{https://github.com/altsoph/RusNor/blob/main/vocabulary/rusnor.v0.1.json}}.
 
\section{Linguistic Discovery Assistant} \label{sec:discovery}

Using the constructed vocabulary as input we developed a Linguistic Discovery Assistant. 
We propose a pipeline for hypothesis generation that leverages a large language model (LLM) as an exploratory tool for analyzing Russenorsk linguistic phenomena. 

\subsection{Agentic Pipeline}
Let us provide an overview of the core agentic steps here. We publish the prompts, the answers, the generate hypotheses and further material regarding to facilitate further work and reproducibility\footnote{\url{https://github.com/altsoph/RusNor/}}.

\paragraph{(1) Lexicon and Origin Hypotheses.} We first supply the LLM with our compiled Russenorsk dictionary (Section~\ref{sec:voc}), prompting it to propose theories regarding the source language of each borrowed term. These requests explicitly ask the model to identify whether the etymon stems from Russian, Norwegian, or a third language.

\paragraph{(2) Phonetics and Morphology.} Next, we reuse the same dictionary to elicit hypotheses about how the morphological and phonetic characteristics of the borrowed terms might evolve when integrated into Russenorsk. The model is thus encouraged to consider potential pathways of phonetic simplification, morphological reduction, or other structural transformations characteristic of contact languages.

\paragraph{(3) Grammar and Syntax.} Finally, we provide the LLM with our dictionary entries alongside example sentences (the same sources used for building the dictionary; see Section~\ref{sec:voc}). We prompt the model to infer possible grammatical rules, such as word-order constraints, morphological markers, or the role of function words within Russenorsk.

\subsection{Ablations and Verification} 
To ensure that the system’s outputs are neither memorized content from model pre-training nor purely “hallucinated” responses, we introduce two ablation scenarios: 
\begin{enumerate} 
\item \textbf{Empty Dictionary (Russenorsk Context).} We inform the model that we are studying Russenorsk but provide an empty dictionary. Any generated hypotheses likely stem from pre-trained data. 
\item \textbf{Fictitious Pidgin.} We inform the model that the target language is a non-existent pidgin and again supply no dictionary. Here, any hypotheses should reflect purely inferential or generic reasoning rather than grounded linguistic knowledge. 
\end{enumerate}

By comparing the model’s proposals in these ablation settings with those generated under full-information conditions, we can isolate the role of the dictionary in shaping its hypotheses. Lastly, we juxtapose the system’s final output with documented linguistic insights from over a century of Russenorsk research. Although historical sources are occasionally inconsistent, mapping the model’s hypotheses to these scholarly discussions highlights both convergences and points of contention, offering a systematic approach to modern Russenorsk inquiry.

\section{Properties of Russenorsk}

In this section, we present key syntactic, morphological, phonetic, and lexical features that characterize Russenorsk, drawing on prior research \newcite{perekhvalskaya_1987, davydov_1986, broch, Fuente_2020}. Most of the human-proposed properties are also proposed by the discovery agent. We publish additional agentic hypotheses, yet further verification of those is beyond the scope of the current study\footnote{\url{https://github.com/altsoph/RusNor/tree/main/responses/}}.

\subsection{Grammar and Syntax}

\paragraph{Word Order.} Although Russenorsk exhibits a relatively free word order, it appears to favor a Subject–Object–Verb (SOV) pattern \newcite{perekhvalskaya_1987}. This flexibility is largely influenced by Russian, where free-word order is common \newcite{davydov_1986}.

\paragraph{Minimal Inflectional Morphology.} Russenorsk generally lacks derivational and inflectional morphology, a characteristic feature of contact languages \newcite{perekhvalskaya_1987}. Nouns do not mark case, numberor gender, and verbs lack categories such as person, number, tense, aspect, and voice \newcite{perekhvalskaya_1987}.

\paragraph{Frozen Forms.} Many Russenorsk words correspond to “frozen” morphological forms derived from Russian or Norwegian. These forms are adopted as-is without further inflection \newcite{broch}.

\paragraph{Verb Suffix -om.} One of the most distinctive features of Russenorsk is the frequent -om suffix on verbs, which serves to mark verbal categories \newcite{perekhvalskaya_1987, Fuente_2020}. For many researchers, this morphological pattern represents Russenorsk’s sole systematic (suffix-based) grammatical form \newcite{broch}.

\paragraph{Absence of Articles.} Russenorsk does not exhibit an article system, closely mirroring Russian’s article-free structure \newcite{broch}.

\paragraph{Universal Preposition [på].} A single multifunctional preposition — commonly realized as [på], [paa], or [pa] — fulfills a wide range of grammatical and semantic roles (e.g., locative, temporal, case-marking) \newcite{perekhvalskaya_1987, broch}. Its etymology combines both Norwegian [på] and Russian [po], reflecting the language’s mixed origins.

\paragraph{Interrogatives and Other Function Words.} 
Russenorsk employs several well-attested strategies for forming questions  \citep{broch}:  
(i) an initial Wh-word (\textit{kak}, \textit{skoro}, etc.);  
(ii) a declarative clause interpreted as a question solely by intonation (e.g.\ \textit{tvoja fisk kupom?});  
(iii) inversion of verb and subject (\textit{vil ju på moja stova …?}); and  
(iv) insertion of the Russian interrogative particle \textit{li} .  

The particle \textit{li}, borrowed from Russian, can stand alone after the focused phrase or appear fused inside lexicalized forms such as \textit{mangeli} / \textit{nogli} ‘how many?’, reflecting its origin in the frequent Russian pattern \textit{mnogo-li} \citep{broch}. Such compounds show that speakers had re-analysed \textit{li} as a productive—if limited—morpheme.

Russenorsk also borrowed the Norwegian discourse particle \textit{kanske} ‘maybe’. Placed sentence-initially, \textit{kanske} functions as a politeness hedge or tentative marker (e.g.\ \textit{kanske tvoja vil glas tšai?} ‘perhaps you’d like a glass of tea’) and is often semantically bleached, bordering on an “empty” pragmatic filler \citep{broch}.

\paragraph{Negation.} Russenorsk expresses negation using forms derived from Russian [njet] and Norwegian [ikke], often interchangeably. In some cases, [njet] may appear as a simple negative marker as well as an indicator of “no” \newcite{broch}.

\paragraph{Lack of Copula.} Russenorsk generally does not employ a copular verb akin to “to be.” This absence is another hallmark of its simplified morphological structure \newcite{perekhvalskaya_1987}.

\subsection{Phonetics and Morphology}

\paragraph{Depalatalization.} As noted in \newcite{Fuente_2020}, Russenorsk, like many other pidgins and creoles, exhibits a strong tendency toward the neutralization of palatal elements. This is exemplified by the loss of the palatal element in the transformation of the Russian -jom into Russenorsk -om.

\paragraph{Convergence of Russian and Norwegian Grammatical Endings.} According to \newcite{Fuente_2020}, the noun marker \textit{-a} is a fusion of elements from Russian and Norwegian, the verb marker \textit{-om} combines the Russian first-person plural and hortative \textit{-jom}, possibly influenced by Cape Dutch pidgin, and the preposition \textit{po} is a blend of the Russian \textit{po} and the Norwegian \textit{på}.

\paragraph{Tendency to add a final syllable.} As noted in \newcite{Hirnsperger2012PidginRussisch}, Russenorsk exhibited a tendency to append an additional syllable to words, despite the fact that monosyllabic words are common in both Russian and Norwegian. For example, the Norwegian word \textit{fisk} (‘fish’) became \textit{fiska} in Russenorsk.

\paragraph{Elimination of front rounded vowels.} \newcite{Belikov2003Russenorsk} provides the following examples, illustrating this tendency, which leads to phonetic simplification. For instance, Norwegian \textit{tønde} (‘barrel’) became Russenorsk \textit{tunna}, and Norwegian \textit{søndag} (‘Sunday’) transformed into Russenorsk \textit{sondag}.

\paragraph{Devoicing of final voiced consonants.} \newcite{Belikov2003Russenorsk} provides the following example, illustrating this tendency: Norwegian \textit{hav} (‘sea’) became Russenorsk \textit{gaf}.

\paragraph{Possible Russian-to-Norwegian phonetic accommodation ([x] → [k]).} \newcite{Belikov2003Russenorsk} states:  
\begin{quote}  
"It is highly probable that the Russians also simplified the phonetics of words of Russian origin, making their pronunciation closer to Norwegian standards (changing, in particular, x to k)."  
\end{quote}

\paragraph{Replacement of glottal [h] by [g].} \newcite{Belikov2003Russenorsk} provides examples of this phenomenon, such as the transformation of Norwegian \textit{halv} (‘half’) into Russenorsk \textit{hal} or \textit{gall}, and Norwegian \textit{hav} (‘sea’) into Russenorsk \textit{gav}.

\paragraph{Splitting up certain Russian consonant clusters.} \newcite{Hirnsperger2012PidginRussisch} notes that certain Russian consonant clusters were split to better align with Norwegian phonetic patterns. For example, Russian \textit{mnogo li} (‘many?’) was adapted in Russenorsk as \textit{nogoli}.

\subsection{Lexicon and Origin Hypotheses}

\paragraph{Fish and Maritime Terms from Norwegian.} Words related to fishing and maritime trade were predominantly borrowed from Norwegian \newcite{davydov_1986} \newcite{davydov_1986}.

\paragraph{Goods of Inland Origin from Russian.} Terms for goods of inland origin, such as flour, bread, and fabric, often have Russian roots \newcite{davydov_1986}.

\paragraph{Greetings and Farewells, Possibly More Frequent from Russian.} \newcite{russenorsk_continuum} notes that all recorded greetings and farewells in Russenorsk contain a Russian element. However, their prevalence may simply reflect the Norwegian provenance of most recorded texts.

\paragraph{Preference for Shorter or Simpler Words.} Since the morphological and phonetic structures of Russenorsk are highly simplified, it predominantly retains lexical items that are the easiest to pronounce and understand, regardless of their original linguistic source \newcite{perekhvalskaya_1987}.

\paragraph{Synonyms of Dual Origin (Russian and Norwegian).} For many concepts in Russenorsk, the existing corpus of texts records the presence of two synonyms, typically one of Russian origin and the other of Norwegian origin \newcite{perekhvalskaya_1987, russenorsk_continuum, broch}.

\paragraph{Single-Origin Synonyms.} \newcite{russenorsk_continuum} states:  
\begin{quote}  
"Single‐source synonym appears to have gone unnoticed in the literature, though synonym groups of this type also occur, albeit not as frequently as multi‐source synonym groups."  
\end{quote}

\paragraph{Merging of Similar Words from Both Source Languages.} When the source languages contain words that are phonetically similar and semantically appropriate, they sometimes merge into a single Russenorsk term—a process referred to as "dual etymology" \newcite{russenorsk_continuum}.

\paragraph{A Small Percentage of Words from Other Languages.} Russenorsk contains a small number of words that likely originate from Dutch, Low German, High German, and English \newcite{russenorsk_continuum}.

\paragraph{Self-Glossing.} Sometimes, synonyms—either from both source languages or even from the same language—are used together to increase the likelihood of correctly conveying the message. Examples include \textit{korosjo dobra} (‘good good’), \textit{stari gammel} (‘old old’), \textit{lita gran nemnožko} (‘a little a little’), and \textit{nokka lite} (‘a little a little’) \newcite{russenorsk_continuum}.

\paragraph{Founder Effect.} According to \newcite{russenorsk_continuum}, certain words may establish themselves as the sole widely used term, possibly due to a founder effect or because one group used them more frequently when initiating contact.

\section{Results of Agentic Discovery}
\label{sec:agentic-rules}

The agent presented in Section \ref{sec:discovery} produced the following key findings:
\begin{itemize}
 \item \textbf{Grammar and Syntax}
    Overall, nearly all hypotheses align with the observed features of Russenorsk. The only contradiction is that the models suggest that Russenorsk exhibits a preference for SVO word order, whereas \newcite{perekhvalskaya_1987} proposes an SOV structure as possibly slightly more characteristic.

    A quick qualitative scan of our corpus suggests that both SVO and SOV clauses occur with comparable frequency, with SVO only slightly more common overall. The persistence of SOV remains noteworthy, however, because that order is not characteristic of either Norwegian or Russian.
    
    \item \textbf{Phonetics and Morphology}
    Overall, all hypotheses of the models align with the features outlined above. The only phonetic characteristic not identified by the models is the replacement of glottal \textit{[h]} by \textit{[g]}.
    \item \textbf{Lexicon and Origin Hypotheses}
    Most lexical hypotheses produced by the models align with the features outlined above. However, the models overlook several attested patterns: (i) the tendency for greetings and farewells to originate primarily from Russian, (ii) the presence of single-origin synonyms, and (iii) the self-glossing strategy in which near-synonymous words are doubled to enhance comprehension. Additionally, two hypotheses—\textit{Fish names from the language of the primary catch zone} and \textit{Preference for words ending in vowels}—lack direct confirmation in the existing literature but appear plausible.
\end{itemize}

Table \ref{tab:hypothesis_evaluation} allows to see which properties of Russenorsk mentioned in the literature were suggested by the generative models prompted along the agentic pipeline described in Section \ref{sec:discovery}. It also shows which of these hypotheses were generated by the agent without any dictionary information, yet with information that we are interested in Russenorsk, as well as by the agent that had to dictionary and no information on which specific pidgin we want to study.

Claude/Sonnet achieves marginally higher recall than o1. At the same time, the union of their outputs provides broader coverage than either model individually, underscoring a useful cumulative effect when both systems are consulted in parallel.

\begin{table*}[h!]
\centering
\renewcommand{\arraystretch}{1.2}
\begin{tabular}{l c c c c}
    \toprule
    \textbf{Hypothesis} & \textbf{Sonnet} & \textbf{o1} & \textbf{Empty Dictionary} & \textbf{Fictitious Pidgin} \\
    \midrule
    \multicolumn{5}{l}{\textbf{Grammar and Syntax}} \\
    Word Order & $\times$ & $\times$ & $\times$ & $\checkmark$ \\
    Minimal Inflectional Morphology & $\checkmark$ & $\checkmark$ & $\checkmark$ & $\times$ \\
    Frozen Forms & $\checkmark$ & $\checkmark$ & $\times$ & $\times$ \\
    Verb Suffix -om & $\checkmark$ & $\checkmark$ & $\times$ & $\times$ \\
    Absence of Articles & $\checkmark$ & $\checkmark$ & $\times$ & $\checkmark$ \\
    Universal Preposition [på] & $\checkmark$ & $\times$ & $\times$ & $\times$ \\
    Interrogatives and Other Function Words & $\checkmark$ & $\checkmark$ & $\times$ & $\times$ \\
    Negation & $\checkmark$ & $\checkmark$ & $\times$ & $\times$ \\
    Lack of Copula & $\checkmark$ & $\checkmark$ & $\times$ & $\times$ \\
    \midrule
    \multicolumn{5}{l}{\textbf{Phonetics and Morphology}} \\
    Depalatalization & $\checkmark$ & $\checkmark$ & $\times$ & $\times$ \\
    Convergence of Grammatical Endings & $\checkmark$ & $\checkmark$ & $\times$ & $\times$ \\
    Tendency to Add a Final Syllable & $\checkmark$ & $\checkmark$ & $\times$ & $\times$ \\
    Elimination of Front Rounded Vowels & $\checkmark$ & $\checkmark$ & $\times$ & $\times$ \\
    Devoicing of Final Voiced Consonants & $\checkmark$ & $\times$ & $\times$ & $\times$ \\
    Russian-to-Norwegian Phonetic Accommodation & $\checkmark$ & $\times$ & $\times$ & $\times$ \\
    Replacement of Glottal [h] by [g] & $\times$ & $\times$ & $\times$ & $\times$ \\
    Splitting up Russian Consonant Clusters & $\checkmark$ & $\checkmark$ & $\checkmark$ & $\checkmark$ \\
    \midrule
    \multicolumn{5}{l}{\textbf{Lexicon and Origin Hypotheses}} \\
    Fish and Maritime Terms from Norwegian & $\checkmark$ & $\checkmark$ & $\times$ & $\times$ \\
    Goods of Inland Origin from Russian & $\checkmark$ & $\checkmark$ & $\times$ & $\times$ \\
    Greetings and Farewells from Russian & $\times$ & $\times$ & $\times$ & $\times$ \\
    Preference for Shorter or Simpler Words & $\checkmark$ & $\checkmark$ & $\times$ & $\times$ \\
    Synonyms of Dual Origin & $\checkmark$ & $\times$ & $\times$ & $\checkmark$ \\
    Single-Origin Synonyms & $\times$ & $\times$ & $\times$ & $\times$ \\
    Merging of Similar Words from Both Languages & $\checkmark$ & $\times$ & $\times$ & $\times$ \\
    A Small Percentage of Words from Other Languages & $\times$ & $\checkmark$ & $\times$ & $\times$ \\
    Self-Glossing & $\times$ & $\times$ & $\times$ & $\times$ \\
    Founder Effect & $\checkmark$ & $\checkmark$ & $\times$ & $\checkmark$ \\
    \bottomrule
\end{tabular}
\caption{Overview of agentic discovery hypotheses across ablations.}
\label{tab:hypothesis_evaluation}
\end{table*}


\section{"Reconstruction" Translation for Extinct Languages} \label{sec:translation_reconstruction}

The final component of our study demonstrates how the pipeline introduced in Section~\ref{sec:discovery} can be extended to "reconstruct" translations for a contact language no longer in active use. Russenorsk serves here as a case study, illustrating the potential for systematic alignment of parallel text, lexical matching, and informed linguistic hypotheses to recover and interpret historical content. Although this approach is not strictly falsifiable, it offers a practical methodology for scholars working with extinct or endangered languages. The prompts for such a "reconstruction" translation are published online\footnote{\url{https://github.com/altsoph/RusNor/blob/main/tr_exp/tr_test1.rusnor.prompt.tpl}}.

\subsection{Pipeline Overview} 
We integrate each stage of the linguistic discovery workflow into a translation task: 
\begin{enumerate} 
\item \textbf{Dictionary Integration.} We supply the model with the Russenorsk lexicon (Section~\ref{sec:voc}) to anchor the translation in the existing word inventory. 
\item \textbf{Examples and Hypotheses.} We present the example sentences\footnote{\url{https://github.com/altsoph/RusNor/blob/main/vocabulary/sentence_examples.txt}} and the hypotheses generated in Section~\ref{sec:discovery}, covering source-language identification, morphological/phonetic adaptation, and grammatical structure. 
\item \textbf{Source.} We provide segments of original Russenorsk text that require translation into modern Russian or Norwegian.
\item \textbf{Prompted Translation.} The model is instructed to reference all prior information (dictionary entries, morphological/phonetic rules, and grammatical constraints) to produce a Russenorsk translation—preferring known lexemes over inventing new forms. 
\end{enumerate}

\subsection{Instruction} 
The pipeline described in the previous subsection results in a "reconstruction" translation into hypothetical historical Russenorsk. While the exact fidelity of the final output is challenging to verify (due to the language’s extinct status and limited corpus), the method provides a principled approach that combines: 
\begin{itemize} 
\item \textbf{Parallel Alignment:} Aligning segments from Russenorsk source text with corresponding pieces in modern languages. 
\item \textbf{Lexical Matching:} Prioritizing existing dictionary forms to maintain consistency and reduce unnecessary neologisms. 
\item \textbf{Hypothesis-Driven Variation:} Incorporating morphological and phonetic adaptation rules derived from Section~\ref{sec:discovery}. 
\end{itemize}

Here is an example of an obtained "reconstruction" Russenorsk translation for a traditional Pomor tale. For the English version we refer the reader to the Appendix \ref{sec:appendix}.

\begin{quote} 
Stannom starik på staraja kona. Odin ras starik på skoga spaserom på trovva rubom. Gak on vil trea rubom, trea prosom: "Njet rubom moja, saa moja på tvoja ønska dast." Starik prosom rik bli. På stova kom - datsja nova stannom, grot kleba stannom, på diengi bolsjoi. Men staraja kona på klokka sprækom: "Kak moja rik levom, men folka på moja njet ærom!" Saa kona vil starik på makta stannom. Trea dast etta ønska. Men kona njet dobra - på makta stannom drogoj makta. Kona prosom starik på trea sprækom: først på bolsjoi makta, saa på større makta, saa på grot makta, saa på tsar, saa på boga stannom. Trea dast altsamma, men på siste ønska starik på kona på bjørna stannom.
\end{quote}

By synthesizing these elements, our pipeline provides a coherent mechanism for generating Russenorsk translations that reflect plausible lexical and grammatical patterns. While certain inferences remain speculative, the overarching framework illuminates both the linguistic “gaps” in historical texts and the utility of modern LLM-driven tools in bridging them.

\subsubsection{A cautious, metric-based evaluation of reconstructed translation quality}
\label{sec:translation-eval}

To obtain a first, \emph{strictly indicative} estimate of translation quality, we
extracted 35 well-formed sentences for which complete triplets were available in
Russian (\textit{ru}), Norwegian (\textit{no}) and legacy Russenorsk
(\textit{rn}).\footnote{The set is released at
\url{https://github.com/altsoph/RusNor/blob/main/tr_exp/tr_bench.json}.}
All automatic translations were produced with
\texttt{anthropic/claude-3-5-sonnet-20241022}; we used the rule hypotheses (introduced in Section~\ref{sec:discovery}), derived with the same model.
For every sentence we asked the model to produce four directions:
\textit{ru}$\to$\textit{rn}, \textit{no}$\to$\textit{rn},
\textit{rn}$\to$\textit{no} and \textit{rn}$\to$\textit{ru}.

\paragraph{Prompt leakage and ablations}
Because the prompt relies on artefacts—the lexicon, the example sentences, and
the hypotheses—derived largely from the same historical sources as our
35-sentence benchmark, the model gains a potential advantage, further
amplified by fragments it may have memorised during pre-training.  To gauge the
scale of this \emph{prompt leakage}, we run a three-step ablation study,
successively removing (1)~the examples, (2)~the lexicon, and (3)~the rules
while keeping every other setting unchanged.

\paragraph{Automatic metric}
Conventional word-level metrics such as BLEU\,\citep{papineni2002bleu}
are unreliable here because both Russenorsk and Russian exhibit heavy inflection,
optional diacritics, and tokenisation ambiguity.  We therefore restrict
ourselves to the character-level F-score chrF\,\citep{popovic2015chrf},
which correlates well with human judgements under such conditions.
Before scoring, both references and hypotheses are lower-cased and stripped of
diacritics.  All chrF values are computed with the \texttt{sacrebleu}
toolkit\,\citep{post2018sacrebleu} to ensure fully reproducible numbers.

\paragraph{Reference points}
As a lower bound we compute chrF between the original Russenorsk sentences and
their Norwegian counterparts (as they share the alphabet and some part of the lexicon); the score is $\approx17$, which we treat as
“chance-level comprehension.”

For a practical upper bound we use the strongest public OPUS-MT models trained
on modern data.  The official benchmark reports
\textbf{chrF 0.418} for Russian$\to$Norwegian
(\href{https://huggingface.co/Helsinki-NLP/opus-mt-ru-no}{opus-mt-ru-no})\,\citep{opusrunocard}
and \textbf{chrF 0.400} for Norwegian$\to$Russian
(\href{https://huggingface.co/Helsinki-NLP/opus-mt-no-ru}{opus-mt-no-ru})\,\citep{opusnorucard}.
These figures therefore set an optimistic ceiling for automatic translation on
this language pair; our Russenorsk reconstructions remain well below that
threshold.

\begin{table*}[t]
  \centering
  \small
  \caption{chrF scores ($\uparrow$) for the four translation directions under
           incremental ablations.  The first row shows a lower‐bound reference
           obtained by comparing the original Russenorsk sentences with their
           Norwegian counterparts. \textcolor{red}{Red} results demonstrate the leakage of exemple sentences. \textcolor{purple}{Purple} results show the model is well familiar with Norwegian and Russian, and many tokens are transparent cognates.}
  \label{tab:chrF_ablations}
  \begin{tabular}{@{}lcccc@{}}
    \toprule
    \textbf{Prompt contents} &
      \textit{ru}$\to$\textit{rn} &
      \textit{no}$\to$\textit{rn} &
      \textit{rn}$\to$\textit{ru} &
      \textit{rn}$\to$\textit{no}\\
    \midrule
    baseline (rn vs.\ no)          & --   & 17.8 & --   & 17.4\\
    dictionary + examples + rules  & \textcolor{red}{67.7} & \textcolor{red}{56.8} & \textcolor{purple}{37.5} & \textcolor{purple}{46.2} \\
    -- examples                    & \textbf{29.2} & \textbf{32.0} & \textcolor{purple}{38.5} & \textcolor{purple}{45.8} \\
    rules only                     & 22.4 & 22.7 & \textcolor{purple}{33.6} & \textcolor{purple}{38.6} \\
    none (memory test)             & 19.7 & 20.4 & \textcolor{purple}{33.2} & \textcolor{purple}{40.0} \\
    \bottomrule
  \end{tabular}
\end{table*}

\paragraph{Interpretation}

We now interpret the chrF scores summarised in Table~\ref{tab:chrF_ablations}.

\begin{itemize}
  \item The dramatic gap between directions \emph{into} Russenorsk and
        \emph{out of} Russenorsk under the full prompt indicates that the model
        often copies or lightly paraphrases example sentences when they are
        available.
  \item Removing those examples lowers \textit{ru}$\to$\textit{rn} and
        \textit{no}$\to$\textit{rn} by roughly 30–40 chrF points, yet they
        still outperform the 17-chrF baseline, showing that the dictionary and
        rules convey additional, non-trivial signal.
  \item With only high-level rules left, performance drops to
        $\approx22$–$23$~chrF for \textit{ru}/\textit{no}$\to$\textit{rn} but
        remains slightly above baseline, hinting at a limited ability to
        \emph{compose} unseen Russenorsk forms.
  \item In the zero-resource setting (\emph{none}) the model still reaches
        33–40~chrF for \textit{rn}$\to$\textit{ru}/\textit{no}, presumably
        because many tokens are transparent cognates, and the model is well familiar with Norwegian and Russian. Translation
        \emph{into} Russenorsk, however, collapses to near-baseline values,
        confirming that productive generation requires explicit lexical
        support.
\end{itemize}

Overall, the experiment shows incremental, measurable gains from each resource
layer while underscoring the ease with which large language models can
over-fit to illustrative material.  Given the tiny sample and the exploratory
nature of chrF in a pidgin context, these figures should be considered a
\emph{sanity check} rather than a definitive quality assessment.
A proper human evaluation — ideally with experts in Scandinavian contact
languages — might be an essential next step.

\subsection{Outlook for Extinct and Endangered Languages} 

Although demonstrated here with Russenorsk, this translation-reconstruction approach can extend to other contact languages or endangered linguistic varieties with similarly sparse documentation. By systematically integrating dictionary resources, morphosyntactic hypotheses, and historical text samples, researchers can better preserve linguistic heritage and potentially uncover previously overlooked facets of these languages’ structures.

\section{Conclusion}

This paper demonstrates how modern large language models could be used in linguistics of low-resource dead languages and pidgins. We create the first structured dictionary of Russenorsk and make it available under an open-source license. The lexicon primarily consists of borrowed terms from Russian and Norwegian, with occasional influences from other languages. The distribution of terms reflects the functional nature of the pidgin as a trade language. We also publish a linguistic discovery agent and demonstrate its capability to generate viable linguistic hypotheses about Russenorsk. We believe similar ideas could be applied to other dead or low-resource pidgin languages. Finally, we provide the first automated pipeline for "reconstructive" translation into Russenorsk.

Russenorsk æ god språk for folk. Vi snakke russisk, norsk, og litt mer, for handel og vennskap. Det viser hvordan folk kan lage nye språk sammen, lett og rask.

\section*{Limitations}
Given the limited documentation of Russenorsk, the output of an agentic discovery pipeline is highly speculative and needs further investigations that include corpus-based analyses, comparative studies with other pidgins, and investigations into the sociolinguistic contexts of its use. The same limitations restricts the "reconstructed" translation pipeline. We use "reconstructed" in the quotation marks to draw attention of the reader to the fact that the resulting output could hardly be called a translation in the common sense of the word.

\section*{Ethics Statement}

This paper complies with the \href{https://www.aclweb.org/portal/content/acl-code-ethics}{ACL Ethics Policy}. The manual annotation of the obtained results was carried out by the co-authors of the paper, thus we acknowledge the potential for annotator bias, which could impact dataset consistency.

\bibliography{ref}

\appendix
\section{Appendix: Tale in English}
\label{sec:appendix}

Traditional Pomor tale about an old man and a tree. Translation into English:
\begin{quote}  
An old man lived with an old woman. One day the old man went into the forest to chop wood. And when he wanted to cut down another tree, it asked not to be cut down, but in return - to fulfill any wish. The old man decided to ask for wealth. He comes home - the hut has become new, there are many domestic animals, bread will last for a long time, "there is more than enough money." But after a while the old woman began to say: "Even though we live richly, what is the point if people do not respect us!", and that it would be good for the grandfather to become a burgomaster. The tree fulfills this wish as well. But the old woman is again dissatisfied - after all, there is power over the burgomaster too. She suggests that the grandfather ask the tree to make him first a master, then a colonel, a general, a king, and then God himself. The tree fulfills all this, but in response to the last request, it turns the old man and the old woman into bears.
\end{quote}




\end{document}